\documentclass[conference,10pt]{IEEEtran}

\usepackage[dvips]{graphicx}
\usepackage[english]{babel}
\usepackage[latin1]{inputenc}

\usepackage{amsmath}
\usepackage{amssymb}
\usepackage{amsbsy}
\usepackage{url}
\usepackage{pstricks}

\usepackage{cite} 
                        
\begin{document}

\title{A new probabilistic transformation \\
of belief mass assignment}
\author{
\authorblockN{Jean Dezert}
\authorblockA{
ONERA\\
The French Aerospace Lab\\
29 Av. Division Leclerc,\\
92320 Ch\^atillon, France.\\
Email:jean.dezert@onera.fr}\and
\authorblockN{Florentin Smarandache}
\authorblockA{Chair of Math. \& Sciences Dept.\\
University of New Mexico,\\
200 College Road,\\
Gallup, NM 87301, U.S.A.\\
Email: smarand@unm.edu}}

\maketitle

\selectlanguage{english}

\begin{abstract}
In this paper, we propose in Dezert-Smarandache Theory (DSmT) framework, a new probabilistic transformation, called DSmP, in order to build a subjective probability measure from any basic belief assignment defined on any model of the frame of discernment. Several examples are given to show how the DSmP transformation works and we compare it to main existing transformations proposed in the literature so far. We show the advantages of DSmP over classical transformations in term of Probabilistic Information Content (PIC). The direct extension of this transformation for dealing with qualitative belief assignments is also presented.
\end{abstract}


%
\IEEEpeerreviewmaketitle
\section{Introduction and motivation}

In the theories of belief functions, Dempster-Shafer Theory (DST) \cite{Shafer76}, Transferable Belief Model (TBM) \cite{Smets90b}  or DSmT \cite{DSmTBook1,DSmTBook2}, the mapping from the belief to the probability domain is a controversial issue.
 The original purpose of such mappings was to make (hard) decision, but contrariwise to erroneous widespread idea/claim,  this is not the only interest for using such mappings nowadays. Actually the probabilistic transformations of belief mass assignments are very useful in modern multitarget multisensor tracking systems (or in any other systems) where one deals with soft decisions (i.e. where all  possible solutions are kept for state estimation with their likelihoods). For example, in a Multiple Hypotheses Tracker using both kinematical and attribute data, one needs to compute all probabilities values for deriving the likelihoods of data association hypotheses and then mixing them altogether to estimate states of targets. Therefore, it is very relevant to use a mapping which provides a high probabilistic information content (PIC) for expecting better performances. This perfectly justifies the theoretical work proposed in this paper. A classical transformation is the so-called {\it{pignistic probability}} \cite{Smets2005}, denoted $BetP$, which offers a good compromise between the maximum of credibility $Bel$ and the maximum of  plausibility $Pl$ for decision-support. Unfortunately, $BetP$ doesn't provide the highest PIC in general as pointed out by Sudano \cite{Sudano2001,Sudano2002,Sudano2003}. We propose hereafter a new generalized pignistic  transformation, denoted $DSmP$, which is justified by the maximization of the PIC criterion. An extension of this transformation in the qualitative domain is also presented.

\section{Pignistic probabilities}

The basic idea  of the pignistic transformation \cite{Smets90,Smets2005} consists in transferring the positive mass of belief of each non specific element onto the singletons involved in that element split by the cardinality of the proposition when working with normalized basic belief assignments (bba's). The (classical) pignistic probability in TBM framework is given by\footnote{We assume that $m(.)$ is of course a non degenerate bba, i.e. $m(\emptyset)\neq 1$.} $BetP(\emptyset)=0$ and $\forall X \in 2^\Theta \setminus\{\emptyset\}$ by:
\begin{eqnarray}
\label{pignistic}
BetP(X)=\sum_{Y \in 2^\Theta, Y \neq \emptyset} \frac{|X \cap Y|}{|Y|} \frac{m(Y)}{1-m(\emptyset)},
\label{eq:Pig}
\end{eqnarray}
where $2^\Theta$ is the power set of the finite and discrete frame $\Theta$ assuming Shafer's model, i.e. all elements of $\Theta$ are assumed truly exclusive. In Shafer's approach, $m(\emptyset)=0$ and the formula \eqref{pignistic} can be rewritten for any singleton $\theta_{i} \in \Theta$ as
\begin{equation}
\label{pignistic1}
BetP(\theta_{i})=\sum_{\substack{Y \in 2^\Theta \\ \theta_{i} \subseteq Y}} \frac{1}{|Y|} m(Y)=m(\theta_{i}) + \sum_{\substack{Y \in 2^\Theta \\ \theta_{i} \subset Y}} \frac{1}{|Y|} m(Y)
\end{equation}
This transformation has been generalized in DSmT for any regular bba $m(.): G^\Theta \mapsto [0,1]$ (i.e. such that $m(\emptyset)=0$ and $\sum_{X \in G^\Theta} m(X)=1$) and for any model of the frame (free DSm model, hybrid DSm model and Shafer's model as well) \cite{DSmTBook1}. It is given by $BetP(\emptyset)=0$ and $\forall X \in G^\Theta \setminus\{\emptyset\}$ by
\begin{equation}
BetP(X)=\sum_{Y \in G^\Theta}  \frac{\mathcal{C}_{\mathcal{M}}(X\cap Y)}{\mathcal{C}_{\mathcal{M}}(Y)}m(Y)
\label{eq:PigG}
\end{equation}
\noindent
where $G^\Theta$ corresponds to the hyper-power set including all the integrity constraints of the model (if any)\footnote{$G^\Theta=2^\Theta$ if one adopts Shafer's model for $\Theta$ and $G^\Theta=D^\Theta$ (Dedekind's lattice) if one adopts the free DSm model for $\Theta$ \cite{DSmTBook1}.}; $\mathcal{C}_{\mathcal{M}}(Y)$ denotes the DSm cardinal\footnote{$\mathcal{C}_{\mathcal{M}}(Y)$ is the number of parts of $Y$ in the Venn diagram of the model ${\mathcal{M}}$ of the frame $\Theta$ under consideration \cite{DSmTBook1} (Chap. 7).} of the set $Y$.  The formula \eqref{eq:PigG} reduces to \eqref{eq:Pig} when $G^\Theta$ reduces to classical power set $2^\Theta$ when one adopts Shafer's model.

\section{Sudano's probabilities}

Recently, Sudano has proposed interesting alternatives denoted $PrPl$, $PrNPl$, $PraPl$, $PrBel$ and $PrHyb$ to $BetP$, all defined in DST framework \cite{Sudano2006}. Sudano uses different kinds of mappings either proportional to the plausibility, to the normalized plausibility, to all plausibilities, to the belief or a hybrid mapping. $PrPl$ and  $PrBel$ are defined\footnote{For notation convenience and simplicity, we use a different but equivalent notation than the one in \cite{Sudano2006}.} for all $X\neq\emptyset \in \Theta$ by:
\begin{equation}
\label{SudanoPrPl}
PrPl(X)=Pl(X) \cdot \sum_{Y \in 2^\Theta, X \subseteq Y} \frac{m(Y)}{CS[Pl(Y)]}
\end{equation}
\begin{equation}
\label{SudanoPrBel}
PrBel(X)=Bel(X) \cdot \sum_{Y \in 2^\Theta, X \subseteq Y} \frac{m(Y)}{CS[Bel(Y)]} 
\end{equation}
\noindent
where the compound-to-sum of singletons (CS) operator of any function\footnote{For example, $f(.)$ must be replaced by $Pl(.)$ in \eqref{SudanoPrPl} or by $Bel(.)$ in \eqref{SudanoPrBel}.} $f(.)$ is defined by \cite{Sudano2001}:
$$CS[f(Y)]\triangleq\sum_{Y_i \in 2^\Theta, |Y_i|=1, \cup_i Y_i=Y} f(Y_i)$$
\noindent
$PrNPl$, $PraPl$ and $PrHyb$ are given by \cite{Sudano2001,Sudano2006}:
\begin{itemize}
\item a mapping proportional to the {\it{normalized}} plausibility 
\begin{equation}
\label{SudanoPrNPl}
PrNPl(X)=\frac{1}{\Delta} \sum_{Y \in 2^\Theta, Y\cap X \neq \emptyset} m(Y)= \frac{1}{\Delta} \cdot Pl(X)
\end{equation}
\noindent
where $\Delta$ is a normalization factor.
\item a mapping proportional to {\it{all}} plausibilities
\begin{equation}
\label{SudanoPraPl}
PraPl(X)=Bel(X) + \epsilon \cdot Pl(X)
\end{equation}
\noindent
with $\epsilon \triangleq  (1-\sum_{Y\in 2^\Theta}Bel(Y)) / (\sum_{Y\in 2^\Theta}Pl(Y)$.
\item a hybrid transformation
\begin{equation}
\label{SudanoPrHyb}
PrHyb(X)=PraPl(X) \cdot \sum_{\substack{Y \in 2^\Theta \\ X \subseteq Y}} \frac{m(Y)}{CS[PraPl(Y)]} 
\end{equation}
\end{itemize}

\section{Cuzzolin's Intersection Probability}

In 2007, a new transformation has been proposed in \cite{Cuzzolin2007} by Cuzzolin in the framework of DST. From a geometric interpretation of Dempster's rule, an {\it{Intersection Probability}} measure was proposed from the proportional repartition of the Total Non Specific Mass\footnote{i.e. the mass committed to partial and total ignorances, i.e. to disjunctions of elements of the frame.} (TNSM) by each contribution of the non-specific masses involved in it. For notation convenience, we will denote it $CuzzP$ in the sequel. $CuzzP(.)$ is defined on any finite and discrete frame $\Theta=\{\theta_1,\ldots,\theta_n\}$, $n\geq 2$, satisfying Shafer's model, by
\begin{equation}
\label{CuzzP}
CuzzP(\theta_i)=m(\theta_i) + \frac{\Delta(\theta_i)}{\sum_{j=1}^n \Delta(\theta_j)}\times TNSM
\end{equation}
\noindent
with $\Delta(\theta_i)\triangleq Pl(\theta_i) - m(\theta_i)$ and 
\begin{equation}
\label{TNSMDefinition}
TNSM=1 - \sum_{j=1}^n m(\theta_j)=\sum_{A \in 2^\Theta, |A|>1} m(A)
\end{equation}

\noindent
$CuzzP$ is however not appealing for the following reasons: 
\begin{enumerate}
\item Although \eqref{CuzzP} does not include explicitly Dempster's rule, its geometrical justification  \cite{Cuzzolin2007,Cuzzolin2008} is strongly conditioned by the acceptance of Dempster's rule as the fusion operator for belief functions. This is a dogmatic point of view we disagree with since it has been recognized since many years by different experts of AI community, that other fusion rules can offer better performances, especially for cases where high conflicting sources are involved.
\item Some parts of the masses of partial ignorance, say $A$, involved in the TNSM, are also transferred to singletons, say $\theta_i \in \Theta$ which are not included in $A$ (i.e. such that $\{\theta_i\}\cap A=\emptyset$). Such transfer  is not good and does not make sense in our point of view. 
To be more clear, let's take $\Theta=\{A,B,C\}$ and $m(.)$ defined on its power set with all masses strictly positive. In that case, $m(A\cup B)>0$ does count in TNSM and thus it is a bit redistributed back to $C$ with the ratio $\frac{\Delta(C)}{\Delta(A)+ \Delta(B) + \Delta(C)}$ through $TNSM>0$. There is no solid reason for committing partially $m(A\cup B)$ to $C$ since, only $A$ and $B$ are involved in that partial ignorance. Similar remark holds for the partial redistribution of $m(A\cup C)>0$.
\item $CuzzP$ is not defined when $m(.)$ is a probabilistic mass because one gets $0/0$ indetermination. This remark is important only from the mathematical point of view.
\end{enumerate}

 \section{A new generalized pignistic transformation}
Our new mapping, denoted $DSmP$ is straight, different from Sudano's and Cuzzolin's mappings which are more refined but less interesting in our opinions than what we present here. The basic idea of $DSmP$ consists in a new way of proportionalizations of the mass of  each partial ignorance such as $A_1\cup A_2$ or $A_1\cup (A_2\cap A_3)$  or $(A_1\cap A_2)\cup (A_3\cap A_4)$, etc. and the mass of the total ignorance $A_1\cup A_2\cup \ldots \cup A_n$, to the elements involved in the ignorances. This new transformation takes into account both the values of the masses and the cardinality of elements in the proportional redistribution process. We first present the general formula for this new transformation and the numerical examples and comparisons with respect to other transformations are given in next sections.

\subsection{The DSmP formula}

Let's consider a discrete frame $\Theta$ with a given model (free DSm model, hybrid DSm model or Shafer's model), the $DSmP$ mapping is defined by\footnote{The formulation of \eqref{eq:DSmP} for the case of singletons $\theta_{i}$ of $\Theta$ is given in \cite{DSmTBook3}.} $DSmP_{\epsilon}(\emptyset)=0$ and $\forall X \in G^\Theta \setminus \{\emptyset\}$ by
\begin{equation}
DSmP_{\epsilon}(X)=\sum_{Y \in G^\Theta}  \frac{\displaystyle\sum_{\substack{Z \subseteq X\cap Y \\ \mathcal{C}(Z)=1}} m(Z) + \epsilon\cdot \mathcal{C}(X\cap Y)}{\displaystyle\sum_{\substack{Z \subseteq Y \\ \mathcal{C}(Z)=1}} m(Z) + \epsilon\cdot \mathcal{C}(Y)}m(Y)
\label{eq:DSmP}
\end{equation}
\noindent
where $\epsilon\geq 0$ is a tuning parameter and $G^\Theta$ corresponds to the hyper-power set including eventually all the integrity constraints (if any) of the model $\mathcal{M}$; $\mathcal{C}(X\cap Y)$ and $\mathcal{C}(Y)$ denote the DSm cardinals\footnote{We have omitted the index of the model $\mathcal{M}$ for notation convenience.} of the sets $X\cap Y$ and $Y$ respectively. $\epsilon$ allows to reach the maximum PIC value of the approximation of $m(.)$ into a subjective probability measure. The smaller $\epsilon$, the better/bigger PIC value. In some particular degenerate cases however, the $DSmP_{\epsilon=0}$ values cannot be derived, but the $DSmP_{\epsilon>0}$ values can however always be derived by choosing $\epsilon$ as a very small positive number, say $\epsilon=1/1000$ for example in order to be as close as we want to the maximum of the PIC (see next sections for details and examples). 
When $\epsilon=1$ and when the masses of all elements $Z$ having $\mathcal{C}(Z)=1$ are zero, \eqref{eq:DSmP} reduces to \eqref{eq:PigG}, i.e. $DSmP_{\epsilon=1}= BetP$. The passage from a free DSm model to a Shafer's model
involves the passage from a structure to another one, and the cardinals change as well in the formula \eqref{eq:DSmP}. 

\subsection{Advantages of DSmP}

$DSmP$ works for all models (free, hybrid and Shafer's). In order to apply classical $BetP$, $CuzzP$ or Sudano's mappings, we need at first to refine the frame (on the cases when it is possible!) in order to work with Shafer's model, and then apply their formulas. In the case where refinement makes sense, then one can apply the other subjective probabilities on the refined frame.  $DSmP$ works on the refined frame as well and gives the same result as it does on the non-refined frame. Thus $DSmP$ with $\epsilon>0$ works on any models and so is very general and appealing. It is a combination of $PrBel$ and $BetP$. $PrBel$ performs a redistribution of an ignorance mass to the singletons involved in that ignorance proportionally with respect to the singleton masses. While $BetP$ also does a redistribution of an ignorance mass to the singletons involved in that ignorance but proportionally with respect to the singleton cardinals. $PrBel$ does not work when the masses of all singletons involved in an ignorance are null since it gives the indetermination 0/0; and in the case when at least one singleton mass involved in an ignorance is zero, that singleton does not receive any mass from the distribution even if it was involved in an ignorance, which is not fair/good. 
So, $DSmP$ solves the $PrBel$ problem by doing a redistribution of the ignorance mass with respect to both the singleton masses and the singletons' cardinals in the same time. Now, if all masses of singletons involved in all ignorances are different from zero, then we can take $\epsilon=0$, and $DSmP$ coincides with $PrBel$ and both of them give the best result, i.e. the best PIC value. 
$PrNPl$ is not satisfactory since it yields to an abnormal behavior. Indeed, in any model, when a bba $m(.)$ is transformed into a probability, normally (we mean it is logically that) the masses of ignorances are transferred to the masses of elements of cardinal 1 (in Shafer's model these elements are singletons). Thus, the resulting probability of an element whose cardinal is 1 should be greater than or equal to the mass of that element. I. e. if $A$ in $G^\Theta$ and $\mathcal{C}(A)=1$, then $P(A) \geq m(A)$ for any probability transformation $P(.)$. This legitimate property is not satisfied by $PrNPl$, since for example if we consider $\Theta=\{A,B,C\}$ and $m(A)=0.2$, $m(B)=m(C)=0$ and $m(B\cup C)=0.8$, one obtains $PrNPl(A)=0.1112  <  m(A)=0.2$. So it is abnormal that singleton $A$ looses mass when $m(.)$ is transformed into a subjective probability. 

In summary, $DSmP$ does an 'improvement' of all Sudano, Cuzzolin, and BetP formulas,
in the sense that $DSmP$ mathematically makes a more accurate redistribution of the ignorance masses to the singletons involved in ignorances. $DSmP$ and $BetP$ work in both theories: DST (= Shafer's model) and DSmT (= free or hybrid models) as well. In order to use Sudano's and Cuzzolin's in DSmT models, we have to refine the frame (see Example 5).

 \section{ The Probabilistic Information Content (PIC)}

Following Sudano's approach \cite{Sudano2001,Sudano2002,Sudano2006}, we adopt the Probabilistic Information Content (PIC) criterion as a metric depicting the strength of a critical decision by a specific probability distribution. It is an essential measure in any threshold-driven automated decision system. The PIC is the dual of the normalized Shannon entropy. A PIC value of one indicates the total knowledge to make a correct decision (one hypothesis has a probability value of one and the rest of zero). A PIC value of zero indicates that the knowledge to make a correct decision does not exist (all the hypotheses have an equal probability value), i.e. one has the maximal entropy. The PIC is used in our analysis to sort the performances of the different pignistic transformations through several numerical examples. We first recall what Shannon entropy and PIC measure are and their tight relationship.

\subsection{Shannon entropy}

Shannon entropy, usually expressed in bits (binary digits), of a probability measure $P\{.\}$ over a discrete finite set $\Theta=\{\theta_1,\ldots,\theta_n\}$ is defined by\footnote{with common convention $0\log_2 0 = 0$.} \cite{Shannon48}:
\begin{equation}
\label{ShannonEntropy}
H(P)\triangleq - \sum_{i=1}^{n} P\{\theta_i\} \log_2(P\{\theta_i\})
\end{equation}
$H(P)$ is maximal for the uniform probability distribution over $\Theta$, i.e. when $P\{\theta_i\}=1/n$ for $i=1,2,\ldots, n$. In that case, one gets $H(P)=H_{\max} =  - \sum_{i=1}^{n} \frac{1}{n} \log_2(\frac{1}{n})= \log_2(n)$. $H(P)$ is minimal for a totally {\it{deterministic}} probability, i.e. for any $P\{.\}$ such that $P\{\theta_i\}=1$ for some $i\in\{1,2,\ldots,n\}$ and $P\{\theta_j\}=0$ for $j\neq i$. $H(P)$ measures the randomness carried by any discrete probability $P\{.\}$. 

\subsection{The PIC metric}
\label{PIC}
The Probabilistic Information Content (PIC) of a probability measure $P\{.\}$ associated with a probabilistic source over a discrete finite set $\Theta=\{\theta_1,\ldots,\theta_n\}$ is defined by \cite{Sudano2002}:
\begin{equation}
\label{PIC}
PIC(P)=1 + \frac{1}{H_{\max}} \cdot \sum_{i=1}^{n} P\{\theta_i\} \log_2(P\{\theta_i\})
\end{equation}
The PIC is nothing but the dual of the normalized Shannon entropy and thus is actually unit less. $PIC(P)$ takes its values in $[0,1]$. $PIC(P)$ is maximum, i.e. $PIC_{\max}=1$ with any {\it{deterministic}} probability and it is minimum, i.e. $PIC_{\min}=0$, with the uniform probability over the frame $\Theta$. The simple relationships between $H(P)$ and $PIC(P)$ are $PIC(P)=1 - (H(P)/H_{\max})$ and $H(P)=H_{\max}\cdot (1-PIC(P))$.

\section{Examples and comparisons on a 2D frame}

Due to the space limitation constraint, all details of derivations are voluntarily omitted here but they will appear in \cite{DSmTBook3}. In this section, we work with the 2D frame $\Theta=\{A,B\}$.

 \subsection{Example 1 (Shafer's model and a general source)}
\label{Example1}

Since one assumes Shafer's model, $G^\Theta=2^\Theta=\{\emptyset,A,B,A\cup B\}$. The non-Bayesian quantitative belief mass is given in Table \ref{MassExample1}. Table \ref{TableExample1} presents the results of the different mappings and their PIC sorted by increasing order. One sees that $DSmP_{\epsilon\rightarrow 0}$ provides same result as $PrBel$ and $PIC(DSmP_{\epsilon\rightarrow 0})$ is greater than the PIC values obtained with $PrNPL$, $BetP$, $CuzzP$, $PrPl$ and $PraPl$.
 \begin{table}[!h]
\centering
 \begin{tabular}{|l|c|c|c|}
    \hline
              & $A$ & $B$ & $A\cup B$ \\
    \hline
$m(.)$  & 0.3 & 0.1 & 0.6\\
    \hline
  \end{tabular}
   \caption{Quantitative inputs for example 1}
   \label{MassExample1}
\end{table}
\vspace{-0.7cm}
\begin{table}[!h]
\centering
 \begin{tabular}{|l|c|c||c|}
    \hline
              & $A$ & $B$  & $PIC(.)$ \\
    \hline
  $PrNPl(.)$   & 0.5625 & 0.4375 & 0.0113 \\
   $BetP(.)$     & 0.6000 & 0.4000 &  0.0291  \\
   $CuzzP(.)$  & 0.6000 & 0.4000 & 0.0291 \\
   $PrPl(.)$      & 0.6375 & 0.3625 & 0.0553 \\
    $PraPl(.)$    & 0.6375 & 0.3625 & 0.0553 \\
    $PrHyb(.)$    & 0.6825 & 0.3175 & 0.0984 \\
   $DSmP_{\epsilon=0.001}(.)$   & 0.7492 & 0.2508 & 0.1875 \\
  $PrBel(.)$   & 0.7500 & 0.2500 & 0.1887 \\
   $DSmP_{\epsilon=0}(.)$   & 0.7500 & 0.2500 & 0.1887\\
  \hline
  \end{tabular}
  \caption{Results for example 1.}
\label{TableExample1}
\end{table}
\vspace{-0.5cm}

 \subsection{Example 2 (Shafer's model and the totally ignorant source)}
\label{Example2}

Let's assume Shafer's model and the vacuous bba characterizing the totally ignorant source, i.e. $m(A\cup B)=1$. It can be verified that all mappings coincide with the uniform probability measure over singletons of $\Theta$, except $PrBel$ which is mathematically not defined in that case. This result can be easily proved for any size of the frame $\Theta$ with $|\Theta|>2$.

 \subsection{Example 3 (Shafer's model and a probabilistic source)}
\label{Example3}

Let's assume Shafer's model and let's see what happens when applying all the transformations on a probabilistic source\footnote{This has obviously no practical interest since the source already provides a probability measure, nevertheless this is very interesting to see the theoretical behavior of the transformations in such case.} which commits a belief mass only to singletons of $2^\Theta$, i.e. a Bayesian mass \cite{Shafer76}. It is intuitively expected that all transformations are idempotent when dealing with probabilistic sources, since actually there is no reason/need to modify $m(.)$ (the input mass) to obtain a new subjective probability measure since $Bel(.)$ associated with $m(.)$ is already a probability measure.  So if we consider for example the uniform Bayesian mass defined by $m_u(A)=m_u(B)=1/2$, it is very easy to verify in this case, that almost all transformations coincide with the (probabilistic) input mass as expected, so that the idempotency property is satisfied. Only Cuzzolin's transformation fails to satisfy this property because in $CuzzP(.)$ formula \eqref{CuzzP} one gets $0/0$ indeterminacy since all $\Delta(.)=0$ in \eqref{CuzzP}. This remark is valid whatever the dimension of the frame $\Theta$ is, and for any Bayesian mass (not only for uniform belief mass). 

 \subsection{Example 4 (Shafer's model and non-Bayesian mass)}
\label{Example4}
Let's assume Shafer's model and the non-Bayesian mass (more precisely the simple support mass) given in Table \ref{MassExample4}. We summarize in Table \ref{TableExample4}, the results obtained with all transformations. One sees that $PIC(DSmP_{\epsilon\rightarrow 0})$ is maximum among all PIC values. $PrBel(.)$ does not work correctly since it can not have a division by zero. We use NaN acronym here standing for {\it{Not a Number}}\footnote{we could also use the standard "N/A" standing for "does not apply".}; even overcoming it\footnote{since the {\it{direct}} derivation of $PrBel(B)$ cannot be done from the formula \eqref{SudanoPrBel} because of the undefined form $0/0$, we could however force it to $PrBel(B)=0$ since $PrBel(B)=1-PrBel(A)=1-1=0$, and consequently we indirectly take $PIC(PrBel)=1$.}, $PrBel$ does not do a fair redistribution of the ignorance $m(A\cup B)=0.6$ because $B$ does not receive anything from the mass 0.6, although $B$ is involved in the ignorance $A\cup B$. All $m(A\cup B)=0.6$ was unfairly redistributed to $A$ only.
 \begin{table}[!h]
\centering
 \begin{tabular}{|l|c|c|c|}
    \hline
              & $A$ & $B$ & $A\cup B$ \\
    \hline
$m(.)$  & 0.4 & 0 & 0.6\\
    \hline
  \end{tabular}
  \caption{Quantitative inputs for example 4}
\label{MassExample4}
\end{table}
\vspace{-0.7cm}
\begin{table}[!h]
\centering
\begin{tabular}{|l|c|c||c|}
    \hline
              & $A$ & $B$  & $PIC(.)$ \\
    \hline
    $PrBel(.)$   & 1 & \bf{NaN} &  \bf{NaN}\\
   $PrNPl(.)$   & 0.6250 & 0.3750 &  0.0455 \\
   $BetP(.)$     & 0.7000 & 0.3000 & 0.1187 \\
   $CuzzP(.)$  & 0.7000 & 0.3000 & 0.1187 \\
   $PrPl(.)$      & 0.7750 & 0.2250 & 0.2308 \\
   $PraPl(.)$    & 0.7750 & 0.2250 & 0.2308 \\
   $PrHyb(.)$    & 0.8650 & 0.1350 & 0.4291 \\
  $DSmP_{\epsilon=0.001}(.)$  & 0.9985 & 0.0015 & 0.9838 \\
  $DSmP_{\epsilon=0}(.)$  & 1 & 0 & 1 \\
   \hline
  \end{tabular}
   \caption{Results for example 4.}
\label{TableExample4}
\end{table}
\vspace{-0.7cm}

The best result is an {\it{adequate probability}}, not {\it{the biggest PIC}} in this case.
This is because $P(B)$ deserves to receive some mass from $m(A\cup B)$, so the most correct result is done by $DSmP_{\epsilon=0.001}$ in Table \ref{TableExample4} (of course we can choose any other very small positive value for $\epsilon$ if we want). Always when a singleton whose mass is zero, but it is involved in an ignorance whose mass is not zero, then $\epsilon$ (in $DSmP$ formula \eqref{eq:DSmP}) should be different from zero.

  \subsection{Example 5 (Free DSm model)}
 \label{Example5}

Let's assume the free DSm model (i.e. $A\cap B\neq \emptyset$) and the generalized mass given in Table \ref{MassExample5}. In the case of free-DSm (or hybrid DSm) models, the pignistic probability and the DSmP can be derived directly from $m(.)$ without the refinement of the frame $\Theta$ whereas Sudano's and Cuzzolin's probabilities cannot be derived directly from their formulas \eqref{SudanoPrPl}-\eqref{CuzzP} for such models. However, they can be obtained indirectly after a refinement of the frame $\Theta$ into $\Theta^{\text{ref}}$ which satisfies Shafer's model. More precisely, instead of working directly on the 2D frame $\Theta=\{A,B\}$ with $m(.)$ given in Table \ref{MassExample5}, we need to work on the 3D frame $\Theta^{\text{ref}}=\{A'\triangleq A\setminus \{A\cap B\}, B'\triangleq B\setminus \{A\cap B\},C'\triangleq A\cap B\}$ satisfying Shafer's model with the equivalent bba $m(.)$ defined as in Table \ref{TableExample5refined}. The results are then given in Table \ref{TableExample5}. One sees that $PIC(DSmP_{\epsilon\rightarrow 0})$ is the maximum value. $PrBel$ does not work correctly because it cannot be directly evaluated for $A$ and $B$ since the underlying $PrBel(A')$ and $PrBel(B')$ are mathematically undefined in such case. If one works on the {\it{refined frame}} $\Theta^{\text{ref}}$ and one applies the $DSmP$ mapping of the bba $m(.)$ defined in Table \ref{TableExample5refined}, one obtains naturally the same results for $DSmP$ as those given in table \ref{TableExample5}. Of course the results of $BetP$ in Table \ref{TableExample5} are the same using directly the formula \eqref{eq:PigG} as those using \eqref{eq:Pig} on $\Theta^{\text{ref}}$. The verification is left to the reader.
 \begin{table}[!h!]
\centering
 \begin{tabular}{|l|c|c|c|c|}
    \hline
              & $A\cap B$ & $A$& $B$ & $A\cup B$ \\
    \hline
$m(.)$  & 0.4 & 0.2 & 0.1 & 0.3\\
    \hline
  \end{tabular}
  \caption{Quantitative inputs for example 5}
\label{MassExample5}
\end{table}
\vspace{-0.7cm}
\begin{table}[!h!]
\centering
 \begin{tabular}{|l|c|c|c|c|}
    \hline
             & $C'$ & $A'\cup C'$& $B'\cup C'$ & $A'\cup B' \cup C'$ \\
    \hline
$m(.)$  & 0.4 & 0.2 & 0.1 & 0.3\\
    \hline
  \end{tabular}
  \caption{Quantitative inputs on the refined frame $\Theta^{\text{ref}}$}
\label{TableExample5refined}
\end{table}
\vspace{-0.7cm}
 \begin{table}[h]
\centering
\begin{tabular}{|l|c|c|c||c|}
    \hline
                            & $A$ & $B$  & $A\cap B$ & $PIC(.)$ \\
 \hline
  $PrNPl(.)$   & 0.7895 & 0.7368 &  0.5263 & 0.0741\\
  $CuzzP(.)$  & 0.8400 & 0.8000 & 0.6400 &  0.1801\\
  $BetP(.)$     & 0.8500 & 0.8000 & 0.6500 & 0.1931\\
  $PraPl(.)$    & 0.8736 & 0.8421 & 0.7157 &  0.2789\\
  $PrPl(.)$      & 0.9083 & 0.8544 & 0.7627 &  0.3570\\
  $PrHyb(.)$   & 0.9471 & 0.9165 & 0.8636 & 0.5544\\
   $DSmP_{\epsilon=0.001}(.)$   & 0.9990 & 0.9988 & 0.9978  & 0.9842\\
$PrBel(.)$   & \bf{NaN} & \bf{NaN} & 1  & 1\\
   $DSmP_{\epsilon=0}(.)$   & 1 & 1 & 1  & 1\\
 \hline
  \end{tabular}
 \caption{Results for example 5.}
\label{TableExample5}
\end{table}
\vspace{-1cm}

\section{Examples on a 3D frame}

We work hereafter on the 3D frame $\Theta=\{A,B,C\}$.

\subsection{Example 6 (Shafer's model and a non-Bayesian mass)}
\label{Example6}
This example is drawn from \cite{Sudano2006}.  Let's  assume Shafer's model and the non-Bayesian belief mass given by $m(A)=0.35$, $m(B)=0.25$, $m(C)=0.02$, $m(A\cup B)=0.20$, $m(A\cup C)=0.07$, $m(B\cup C)=0.05$ and $m(A\cup B\cup C)=0.06$. The results of the mappings are given in Table \ref{TableExample6}. One sees that $DSmP_{\epsilon\rightarrow 0}$ provides the same result as $PrBel$ which corresponds here to the best result in term of PIC metric.
\begin{table}[!h]
\centering
\begin{tabular}{|l|c|c|c||c|}
    \hline
     & $A$ & $B$  & $C$ & $PIC(.)$ \\
    \hline
   $PrNPl(.)$   & 0.4722 & 0.3889 &  0.1389 & 0.0936\\
   $CuzzP(.)$  & 0.5029 & 0.3937 & 0.1034 &  0.1377\\
   $BetP(.)$     & 0.5050 & 0.3950 & 0.1000 &  0.1424\\
   $PraPl(.)$    & 0.5294 & 0.3978 & 0.0728 &  0.1861\\
   $PrPl(.)$      & 0.5421 & 0.4005 & 0.0574 &  0.2149\\
   $PrHyb(.)$   & 0.5575 & 0.4019 & 0.0406 & 0.2517\\
   $DSmP_{\epsilon =0.001}(.)$      &  0.5665 & 0.4037 &  0.0298 &  0.2783\\
   $PrBel(.)$   & 0.5668 & 0.4038 & 0.0294  & 0.2793\\
   $DSmP_{\epsilon =0}(.)$   & 0.5668 & 0.4038 & 0.0294  & 0.2793\\
   \hline
  \end{tabular}
  \caption{Results for example 6.}
\label{TableExample6}
\end{table}

\subsection{Example 7 (Shafer's model and a non-Bayesian mass)}
\label{Example7}
Let's assume Shafer's model and change a bit the non-Bayesian input mass by taking $m(A)=0.10$, $m(B)=0$, $m(C)=0.20$, $m(A\cup B)=0.30$, $m(A\cup C)=0.10$, $m(B\cup C)=0$ and $m(A\cup B\cup C)=0.30$. The results of the mappings are given in Table \ref{TableExample7}. One sees that  $DSmP_{\epsilon \rightarrow 0}$ provides the best PIC value than all other mappings since $PrBel$ is mathematically undefined. If one takes artificially $PrBel(B)=0$,  one gets the same result as with $DSmP_{\epsilon \rightarrow 0}$.
 \begin{table}[!htb]
\centering
\begin{tabular}{|l|c|c|c||c|}
    \hline
                            & $A$ & $B$  & $C$ & $PIC(.)$ \\
    \hline
       $PrBel(.)$   & 0.5333 & {\bf{NaN}} & 0.4667  & {\bf{NaN}}\\
      $PrNPl(.)$   & 0.4000 & 0.3000&  0.3000 & 0.0088\\
      $CuzzP(.)$  & 0.3880 & 0.2470 & 0.3650 &  0.0163\\
      $BetP(.)$     & 0.4000 & 0.2500 & 0.3500 &  0.0164\\
      $PraPl(.)$    & 0.3800 & 0.2100 & 0.4100 &  0.0342\\
      $PrPl(.)$      & 0.4486 & 0.2186 & 0.3328 &  0.0368\\
      $PrHyb(.)$   & 0.4553 & 0.1698 & 0.3749 & 0.0650\\
     $DSmP_{\epsilon =0.001}(.)$      & 0.5305 & 0.0039 &  0.4656 &  0.3500\\
   \hline
  \end{tabular}
  \caption{Results for example 7.}
\label{TableExample7}
\vspace{-0.5cm}
\end{table}

\subsection{Example 8 (Hybrid DSm model)}
\label{Example8}

\begin{figure}[!h]
\begin{center}
{\tt \setlength{\unitlength}{1pt}
\begin{picture}(90,90)
\thinlines    
\put(40,60){\circle{40}}
\put(60,60){\circle{40}}
\put(50,10){\circle{40}}
\put(15,84){\vector(1,-1){10}}
\put(7,84){$A$}
\put(84,84){\vector(-1,-1){10}}
\put(85,84){$B$}
\put(85,10){\vector(-1,0){15}}
\put(87,7){$C$}
\put(45,59){$D'$}
\put(47,7){$C'$}
\put(65,59){$B'$}
\put(28,59){$A'$}
\end{picture}}
\end{center}
\caption{Refined 3D frame for example 8}
\label{FigHybridDSmModelRefined}
\end{figure}
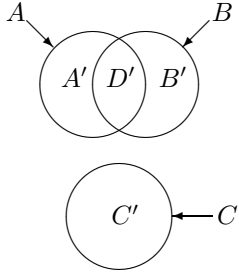

We consider the hybrid DSm model in which all intersections of elements of $\Theta$ are empty, but $A\cap B$. In this case, $G^\Theta$ reduces to 9 elements $\{\emptyset, A\cap B, A, B, C, A\cup B, A\cup C, B\cup C, A\cup B \cup C\}$. The input masses of focal elements are given by $m(A\cap B)=0.20$, $m(A)=0.10$, $m(C)=0.20$, $m(A\cup B)=0.30$, $m(A\cup C)=0.10$, and $m(A\cup B\cup C)=0.10$. In order to apply Sudano's and Cuzzolin's mappings, we need to work on the refined frame $\Theta^{\text{ref}}$ with Shafer's model as depicted on Figure \ref{FigHybridDSmModelRefined} and masses given in the Table \ref{MassExample8refined}. 
\begin{table}[!h]
\centering
\centering
 \begin{tabular}{|l|c|c|c|}
    \hline
     & $D'$ & $A'\cup D'$  & $C'$\\
    \hline
$m(.)$  & 0.2 & 0.1 & 0.2 \\
    \hline
   \hline
     & $A'\cup B'\cup D'$ & $A'\cup C'\cup D'$  & $A'\cup B'\cup C'\cup D'$ \\
    \hline
  $m(.)$  & 0.3 &  0.1 & 0.1\\
    \hline
  \end{tabular}
  \caption{Quantitative inputs on the refined frame for example 8}
\label{MassExample8refined}
\end{table}
\vspace{-0.7cm}

One sees from the Table \ref{TableExample8} that $DSmP_{\epsilon \rightarrow 0}$ provides the best results in term of PIC metric. The refined frame has been defined as:
$\Theta^{\text{ref}}=\{A'\triangleq A\setminus(A\cap B),B'\triangleq B\setminus(A\cap B),C'\triangleq C,D'\triangleq A\cap B\}$ according to Figure \ref{FigHybridDSmModelRefined}.

\clearpage
\newpage

 \begin{table}[!h]
\centering
\begin{tabular}{|l|c|c|c|c||c|}
    \hline
                            & $A'$ & $B'$  & $C'$ & $D'$ & $PIC(.)$ \\
    \hline
       $PrBel(.)$   & {\bf{NaN}} &  {\bf{NaN}} &  0.3000 &  0.7000  & {\bf{NaN}}\\
       $PrNPl(.)$   & 0.2728 & 0.1818 &  0.1818 &  0.3636 & 0.0318 \\
       $CuzzP(.)$  & 0.2000 & 0.1333 & 0.2667 &  0.4000 & 0.0553\\
       $BetP(.)$     &  0.2084 & 0.1250 & 0.2583 &  0.4083 & 0.0607 \\
       $PraPl(.)$    & 0.1636 & 0.1091 & 0.3091 &  0.4182 & 0.0872\\
       $PrPl(.)$      &  0.2035 & 0.0848 & 0.2404 &  0.4713 & 0.1124\\
       $PrHyb(.)$   & 0.1339 & 0.0583 & 0.2656 & 0.5422 & 0.1928\\
       $DSmP_{\epsilon =0.001}(.)$      & 0.0025 & 0.0017 &  0.2996 &  0.6962 & 0.5390\\
   \hline
  \end{tabular}
  \caption{Results for example 8.}
\label{TableExample8}
\end{table}
\vspace{-0.7cm}

\subsection{Example 9 (free DSm model)}
\label{Example9}

We consider the free DSm model depicted on Figure \ref{FigFree3DDSmModel} with the input masses given in Table \ref{MassExample9}. To apply Sudano's and Cuzzolin's mappings, one works on the refined frame $\Theta^\text{ref}=\{A',B',C',D',E',F',G'\}$ where the elements of $\Theta^\text{ref}$ are exclusive (assuming such refinement has a physically sense) according to Figure \ref{FigFree3DDSmModel}. This refinement step is not necessary when using $DSmP$ since it works directly on DSm free model. The PIC values obtained with the different mappings are given in Table \ref{TableExample9}. One sees that $DSmP_{\epsilon \rightarrow 0}$ provides here again the best results in term of PIC. 
\begin{figure}[!h]
\begin{center}
{\tt \setlength{\unitlength}{1pt}
\begin{picture}(90,90)
\thinlines    
\put(40,60){\circle{40}}
\put(60,60){\circle{40}}
\put(50,40){\circle{40}}
\put(15,84){\vector(1,-1){10}}
\put(7,84){$A$}
\put(84,84){\vector(-1,-1){10}}
\put(85,84){$B$}
\put(74,15){\vector(-1,1){10}}
\put(75,10){$C$}
\put(45,64){$D'$}
\put(45,50){$G'$}
\put(47,28){$C'$}
\put(32,43){$E'$}
\put(58,43){$F'$}
\put(65,60){$B'$}
\put(26,60){$A'$}
\end{picture}}
\end{center}
\vspace{-0.5cm}
\caption{Free DSm model for a 3D frame for example 9.}
\label{FigFree3DDSmModel}
\end{figure}
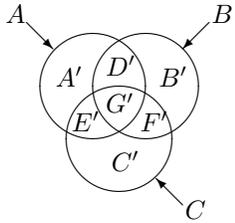
\begin{table}[!h]
\centering
 \begin{tabular}{|l|c|c|c|}
    \hline
     & $A\cap B \cap C$ & $A\cap B$  & $A$\\
    \hline
$m(.)$  & 0.1 & 0.2 & 0.3 \\
    \hline
   \hline
     & $A\cup B$ & $A\cup B \cup C$  &  \\
    \hline
  $m(.)$  & 0.1 &  0.3 &  \\
    \hline
  \end{tabular}
  \caption{Quantitative inputs for example 9}
\label{MassExample9}
\end{table}
\vspace{-0.7cm}
 \begin{table}[!h]
\centering
\begin{tabular}{|l|c|}
    \hline
        Transformations                    & $PIC(.)$ \\
    \hline
        $PrBel(.)$   & {\bf{NaN}} \\
       $PrNPl(.)$   &  0.0414\\
       $CuzzP(.)$  &  0.0621\\
       $PraPl(.)$    & 0.0693\\
       $BetP(.)$     &  0.1176\\
       $PrPl(.)$      &  0.1940\\
       $PrHyb(.)$   & 0.2375 \\
       $DSmP_{\epsilon =0.001}(.)$      & 0.8986\\
   \hline
  \end{tabular}
  \caption{Results for example 9.}
\label{TableExample9}
\end{table}

 \section{Extension of DSmP for qualitative belief}

 \subsection{Qualitative belief assignment $qm(.)$}

In order to compute directly with words (linguistic labels), Smarandache and Dezert have defined in \cite{DSmTBook2} a {\it{qualitative basic belief assignment}} $qm(.)$ as a mapping function from $G^\Theta$ into a set of linguistic labels $L=\{L_0,\tilde{L},L_{n+1}\}$ where $\tilde{L}=\{L_1,\cdots,L_n\}$
is a finite set of linguistic labels and where $n \ge 2$ is an integer. For example, $L_1$ can take the linguistic value ``poor'',
$L_2$ the linguistic value ``good'', etc. $\tilde{L}$ is endowed with a total order relationship $\prec$, so that $L_1 \prec L_2 \prec \cdots \prec L_n $. To work on a true closed linguistic set $L$ under linguistic operators, $\tilde{L}$ is extended with two extreme values $L_0=L_{\min}$ and $ L_{n+1}=L_{\max}$, where $L_0$ corresponds to the minimal
qualitative value and $L_{n+1}$ corresponds to the maximal qualitative value, in such a way that $L_0 \prec L_1 \prec L_2 \prec \cdots \prec L_n  \prec L_{n+1} $, where $\prec$ means inferior to, or less (in quality) than, or smaller than, etc. 
\subsection{Operator on qualitative labels}
\label{subsection:q-operators}
From the extension of the isomorphism between the set of linguistic equidistant labels and a set of numbers in the interval $[0,1]$, one can built exact operators on linguistic labels which makes possible the extension all the quantitative fusion rules and probabilistic transformations into their qualitative counterparts \cite{Li07}.
We briefly remind the main qualitative operators (or $q$-operators for short) on linguistic labels:
\begin{itemize}
\item $q$-addition:
\begin{equation}
L_i+L_j=
\begin{cases}
L_{i+j} & \text{if }  i+j < n+1,\\
L_{n+1}=L_{\max} &  \text{if }  i+j \geq n+1.
\end{cases}
\label{eq:q-addition}
\end{equation}
The $q$-addition is an extension of the addition operator on equidistant labels which is given by $L_i+L_j=\frac{i}{n+1}+ \frac{j}{n+1}=\frac{i+j}{n+1}=L_{i+j}$.
\end{itemize}

\begin{itemize}
\item $q$-subtraction:  
\begin{equation}
 L_i - L_j =
\begin{cases}
L_{i-j} & \text{if} \quad i \geq  j,\\
- L_{j-i} &  \text{if} \quad i <  j.
\end{cases}
\label{eq:qsub}
\end{equation}
\noindent
where $ -L =  \{ -L_1, -L_2, \ldots, -L_n, -L_{n+1} \}$. 
The $q$-subtraction is justified since when $i \geq  j$, one has with equidistant labels $L_i - L_j =  \frac{i}{n+1} - \frac{j}{n+1} =  \frac{i-j}{n+1}$.
\end{itemize}

\begin{itemize}
\item $q$-multiplication\footnote{The $q$-multiplication of two linguistic labels defined here can be extended directly to the multiplication of $n>2$ linguistic labels. For example the product of three linguistic label will be defined as $L_i \cdot L_j  \cdot L_k = L_{[(i\cdot j\cdot k)/(n+1)(n+1)]}$, etc.}:
\begin{equation}
L_i \cdot L_j = L_{[(i\cdot j)/(n+1)]}.
\label{eq:qmult}
\end{equation}
\noindent
where $[x]$ means the closest integer to $x$ (with $[n+0.5]=n+1$,  $\forall n\in \mathbb{N}$). This operator is justified by the approximation of the product of equidistant labels given by $L_i \cdot L_j= \frac{i}{n+1}\cdot  \frac{j}{n+1} = \frac{(i\cdot j)/(n+1)}{n+1}$.
\end{itemize}

\begin{itemize}
\item Scalar multiplication of a linguistic label: Let $a$ be a real number. The multiplication of a linguistic label by a scalar is defined by:
\begin{equation}
a \cdot L_i =\frac{a\cdot i}{n+1} \approx
\begin{cases}
L_{[a\cdot i]} & \text{if} \ [a\cdot i]\geq  0,\\
L_{-[a\cdot i]} & \text{otherwise}.
\end{cases}
\label{eq:sqmult}
\end{equation}
\end{itemize}

\begin{itemize}
\item Division of linguistic labels:
\begin{itemize}
\item[a)]  $q$-division as an internal operator: Let $j\neq 0$, then
\begin{equation}
L_i / L_j  =
\begin{cases}
L_{[(i/j)  \cdot (n+1)]} & \text{if} [(i/j) \cdot (n+1)] < n+1,\\
L_{n+1} & \text{otherwise}.
\end{cases}
\label{eq:sqdiv}
\end{equation}
The first equality in \eqref{eq:sqdiv} is well justified because with equidistant labels, one gets: 
$L_i / L_j  = \frac{i/(n+1)}{j/(n+1)}=\frac{(i/j)\cdot (n+1)}{n+1}\approx L_{[(i/j)  \cdot (n+1)]}$.
\item[b)] Division as an external operator: $\oslash$. Let $j\neq 0$. We define: 
\begin{equation}
L_i \oslash L_j  = i/j.
\label{eq:sqextdiv}
\end{equation}
since for equidistant labels $L_i \oslash L_j = (i/(n+1)) / (j/(n+1)) = i/j$.
\end{itemize}
\end{itemize}

\noindent
{\it{Remark}}: When working with labels, no matter how many operations we have, the best (most accurate) result is obtained if we do only one approximation, and that one should be just at the very end. 

\subsection{More operations with labels}

On the interval $[0,1]$ we consider the labels $L_{i}$, $0\leq i \leq n+1$, $n\geq 0$ such that $L_{i}=i/(n+1)$. But we extend this closed interval to the right and to the left in order to be able to do all needed label operations in any fusion calculation. Therefore $L_{n+2}=\frac{n+2}{n+1}$, $L_{n+3}=\frac{n+3}{n+1}$, \ldots and respectively $L_{-i}=-L_{i}=\frac{-i}{n+1}$, so we get $L_{-1}$, $L_{-2}$, \ldots. In general $L_{i}=i/(n+1)$ for any $i\in \mathbb{Z}=\{\ldots,-2,-1,0,1,2,\ldots\}$ where $\mathbb{Z}$ is the set of all integers. Now we define four more operators involving labels.

\subsubsection{Addition of labels with real scalars}
If $r \in \mathbb{R}$ (the set of real numbers) and $i\in \mathbb{Z}$, then:
\begin{equation}
L_{i} + r= r + L_{i}= L_{[i+ r (n+1)]}
\end{equation}
\noindent
where $[x]$ means the closest integer to $x$. This operator is justified because $L_{i} + r=\frac{i}{n+1}+r=\frac{i+r (n+1)}{n+1}\approx L_{[i+ r (n+1)]}$ and it is needed in the qualitative extension of DSmP formula.

\subsubsection{Subtraction between labels and real scalars}
\begin{equation}
L_{i}- r= L_{[i- r (n+1)]}
\end{equation}
\noindent
because $L_{i} - r=\frac{i}{n+1}-r=\frac{i-r (n+1)}{n+1}\approx L_{[i- r (n+1)]}$ and similarly $r-L_{i}= L_{[r ( n+1) - i]}$ because $r-L_{i} =r-\frac{i}{n+1}=\frac{r (n+1)-i}{n+1}\approx L_{[r (n+1) - i]}$.

\subsubsection{\& {\it{4)}} Powers and roots of labels}
\begin{equation}
(L_{i})^k= L_{[\frac{i^k}{(n+1)^{k-1}}]}
\label{eq:power}
\end{equation}
\noindent
for $k\in  \mathbb{R}$ because $(L_{i})^k= (\frac{i}{n+1})^k=\frac{\frac{i^k}{(n+1)^{k-1}}}{n+1}\approx L_{[\frac{i^k}{(n+1)^{k-1}}]}$.

If $k\in  \mathbb{Q}$, which is the set of fractions (rational numbers), we get the radical operation of labels. Therefore, 
 \begin{equation}
\sqrt[p]{L_{i}}= L_{[\sqrt[p]{i.(n+1)^{p-1}}]}
\end{equation}
\noindent
because we replace $k=1/p$ in the formula \eqref{eq:power}.

\subsection{Quasi-normalization of $qm(.)$}

There is no way to define a normalized $qm(.)$, but a qualitative quasi-normalization \cite{DSmTBook2} is nevertheless possible when considering equidistant linguistic labels because in such case, $qm(X_i) = L_i$, is equivalent to a quantitative mass $m(X_i) = i/(n+1)$ which is normalized if: 
$$\sum_{X\in G^\Theta} m(X)= \sum_{k} i_k/(n+1)=1,$$
\noindent
but this one is equivalent to: 
$$\sum_{X\in G^\Theta} qm(X)= \sum_{k} L_{i_k}=L_{n+1}.$$
\noindent
In this case, we have a {\it{qualitative normalization}}, similar to the (classical) numerical normalization. But, if the labels $L_0$, $L_1$, $L_2$, $\ldots$, $L_n$, $L_{n+1}$ are not equidistant, so the interval $[0, 1]$ cannot be split into equal parts according to the distribution of the labels, then it makes sense to consider a {\it{qualitative quasi-normalization}}, \emph{i.e.} an approximation of the (classical) numerical normalization for the qualitative masses in the same way:
 $$\sum_{X\in G^\Theta} qm(X)=L_{n+1}.$$
\noindent
In general, if we don't know if the labels are equidistant or not, we say that a qualitative mass is quasi-normalized when the above summation holds.

\subsection{Qualitative extension of DSmP}

The qualitative extension of \eqref{eq:DSmP}, denoted $qDSmP(.)$ is given by $qDSmP_{\epsilon}(\emptyset)=0$ and $\forall X \in G^\Theta \setminus \{\emptyset\}$ by
\begin{equation}
qDSmP_{\epsilon}(X)=\sum_{Y \in G^\Theta}  \frac{\displaystyle\sum_{\substack{Z \subseteq X\cap Y \\ \mathcal{C}(Z)=1}} qm(Z) + \epsilon\cdot \mathcal{C}(X\cap Y)}{\displaystyle\sum_{\substack{Z \subseteq Y \\ \mathcal{C}(Z)=1}} qm(Z) + \epsilon\cdot \mathcal{C}(Y)}qm(Y)
\label{eq:qDSmP}
\end{equation}
\noindent
where all operations in \eqref{eq:qDSmP} are referred to labels, that is $q$-operators on linguistic labels defined in \ref{subsection:q-operators} and not classical operators on numbers. In the same manner, due to our construction of labels and qualitative operators, we can transform any quantitative fusion rule (or arithmetic expression) into a qualitative fusion rule (or qualitative expression).

\subsection{Derivation of PIC from qDSmP}

We propose here the derivation of PIC from qualitative DSmP. 
Let's consider a finite space of discrete exclusive events $\Theta=\{\theta_1,\theta_2,\ldots,\theta_M\}$ and a subjective qualitative alike probability measure $qP(.):\Theta \mapsto L=\{L_0,L_1,\ldots,L_n,L_{n+1}\}$. Then one defines the entropy and PIC metrics from $qP(.)$ as

\begin{equation}
\label{qShannonEntropy}
H(qP)\triangleq - \sum_{i=1}^{M} qP\{\theta_i\} \log_2(qP\{\theta_i\})
\end{equation}

\begin{equation}
\label{qPIC}
PIC(qP)=1 + \frac{1}{H_{\max}} \cdot \sum_{i=1}^{M} qP\{\theta_i\} \log_2(qP\{\theta_i\})
\end{equation}

\noindent
where $H_{\max} = \log_2(M)$ and in order to compute the logarithms, one utilized the isomorphism $L_i=i/(n+1)$.

 \section{Example for qualitative DSmP}

Let's consider the frame $\Theta=\{A,B,C\}$ with Shafer's model and the following set of linguistic labels $L=\{L_0,L_1,L_2,L_3,L_4,L_5\}$, with $L_0=L_{\min}$ 
and $L_5=L_{\max}$. Let's consider the following qualitative belief assignment $qm(A)=L_1$, $qm(B\cup C)=L_4$ and $qm(X)=L_0$ for all $X\in 2^\Theta\setminus\{A,B\cup C\}$. $qm(.)$ is quasi-normalized since $\sum_{X\in 2^\Theta} qm(X)=L_5=L_{\max}$. In this example, $qm(B\cup C)=L_4$ is redistributed by $qDSmP_{\epsilon}(.)$ to $B$ and $C$ only, since $B$ and $C$ were involved in the ignorance, proportionally with respect to their cardinals (since their masses are $L_0\equiv 0$). Applying $qDSmP_{\epsilon}(.)$ formula \eqref{eq:qDSmP}, one gets for this example:
$$qDSmP_{\epsilon}(A)=L_1$$
\begin{align*}
qDSmP_{\epsilon}(B)&=\frac{qm(B) + \epsilon\cdot \mathcal{C}(B)}{qm(B) + qm(C) + \epsilon\cdot \mathcal{C}(B\cup C)} qm(B\cup C)\\
& = \frac{L_{0}+ \epsilon\cdot 1}{L_{0}+ L_{0} + \epsilon\cdot 2} \cdot L_4
=\frac{L_{[0+(\epsilon \cdot 1) \cdot 5]}}{L_{[0+0+(\epsilon\cdot 2)\cdot 5]}}\cdot L_4\\
&=\frac{L_{[\epsilon \cdot 5]}}{L_{[\epsilon \cdot 10]}}\cdot L_4=L_{[\frac{5\epsilon}{10\epsilon}\cdot 5]} \cdot L_4=L_{[2.5]} \cdot L_4 \\
&= L_{[ 2.5 \cdot 4 /5]}= L_{[10/5]}=L_2
\end{align*}
Similarly, one gets
\begin{align*}
qDSmP_{\epsilon}(C)&=\frac{qm(C) + \epsilon\cdot \mathcal{C}(C)}{qm(B) + qm(C) + \epsilon\cdot \mathcal{C}(B\cup C)} qm(B\cup C)\\
& = \frac{L_{0}+ \epsilon\cdot 1}{L_{0}+ L_{0} + \epsilon\cdot 2} L_4=L_2
\end{align*}
\noindent
where the index in $[\cdot]$ has been computed at the very end for the best accuracy. Thanks to the isomorphism between labels and numbers, all the properties of operations with numbers are transmitted to the operations with labels.
$qDSmP_{\epsilon}(.)$ is quasi-normalized since $qDSmP_{\epsilon}(A)+qDSmP_{\epsilon}(B)+qDSmP_{\epsilon}(C)$ equals
 $L_1+L_2+L_2=L_5=L_{\max}$. Applying the PIC formula \eqref{qPIC}, one obtains (here $M=\mid \Theta\mid = 3$):
\begin{multline*}
PIC(qDSmP_{\epsilon})=1 + \frac{1}{\log_2 3} (L_1\log_2(L_1) \\
+ L_2\log_2(L_2) +L_2\log_2(L_2)) \approx \frac{1}{5} L_{1}
\end{multline*}
\noindent
where in order to compute the qualitative logarithms, one utilized the isomorphism $L_{i}=\frac{i}{n+1}$.

\section{Conclusions}
\label{conclusion}

Motivated by the necessity to use a better (more informational) probabilistic approximation of belief assignment $m(.)$ for applications involving soft decisions, we have developed a new probabilistic transformation, called $DSmP$,  for approximating $m(.)$ into a subjective probability measure. $DSmP$ provides the maximum of the Probabilistic Information Content (PIC) of the source because it is based on proportional redistribution of partial and total uncertainty masses to elements of cardinal 1 with respect to their corresponding masses and cardinalities. $DSmP$ works directly for any model (Shafer's, hybrid, or free DSm model) of the frame of the problem and the result can be obtained at any level of precision by a tuning positive parameter $\epsilon >0$. $DSmP_{\epsilon=0}$ coincides with Sudano's $PrBel$ transformation for the cases when all masses of singletons involved in ignorances are nonzero. $PrBel$ formula is restricted to work on Shafer's model only while $DSmP_{\epsilon>0}$ is always defined and for any model. We have clearly proved through simple examples that the classical $BetP$ and Cuzzolin's transformations do not perform well in term of PIC criterion. It has been shown also how $DSmP$ can be extended to the qualitative domain to approximate qualitative belief assignments provided by human sources in natural language.

\bibliographystyle{IEEEtran}

\begin{thebibliography}{1}
\providecommand{\url}[1]{#1}
\csname url@rmstyle\endcsname
\providecommand{\newblock}{\relax}
\providecommand{\bibinfo}[2]{#2}
\providecommand\BIBentrySTDinterwordspacing{\spaceskip=0pt\relax}
\providecommand\BIBentryALTinterwordstretchfactor{4}
\providecommand\BIBentryALTinterwordspacing{\spaceskip=\fontdimen2\font plus
\BIBentryALTinterwordstretchfactor\fontdimen3\font minus
  \fontdimen4\font\relax}
\providecommand\BIBforeignlanguage[2]{{
\expandafter\ifx\csname l@#1\endcsname\relax
\else
\language=\csname l@#1\endcsname
\fi
#2}}

\bibitem{Cuzzolin2007} 
F. Cuzzolin, ``On the properties of the Intersection probability'', \emph{submitted to the Annals of Mathematics and AI}, Feb. 2007.

\bibitem{Cuzzolin2008} 
F. Cuzzolin, ``A geometric approach to the theory of  evidence",  \emph{IEEE Transactions on Systems, Man, and Cybernetics}, Part C, 2008 (to appear).{\small{http://perception.inrialpes.fr/people/Cuzzolin/pubs.html}}

\bibitem{Li07} 
X.~Li and X.~Huang and J.~Dezert and F.~Smarandache, ``Enrichment of Qualitative Belief for Reasoning under Uncertainty'', \emph{Proc. of Fusion 2007}, Qu\'ebec, July 2007.

\bibitem{Shafer76} 
G.~Shafer, ``A mathematical theory of evidence'', \emph{Princeton University Press}, 1976.

\bibitem{Shannon48} 
C.E.~Shannon, ``A Mathematical Theory of Communication'', \emph{Bell Syst. Tech. J.}, 27, pp. 379-423 and 623-656, 1948.

\bibitem{DSmTBook1}
F.~Smarandache and  J.~Dezert (Editors), ``Applications and Advances of DSmT for Information Fusion', \emph{American Research Press}, 2004.
{\small{http://www.gallup.unm.edu/{\verb+~+}smarandache/DSmT-book1.pdf}}.

\bibitem{DSmTBook2}
F.~Smarandache and  J.~Dezert (Editors), ``Applications and Advances of DSmT for Information Fusion', Vol.2, \emph{American Research Press}, 2006.
{\small{http://www.gallup.unm.edu/{\verb+~+}smarandache/DSmT-book2.pdf}}.

\bibitem{DSmTBook3}
F.~Smarandache and  J.~Dezert (Editors), ``Applications and Advances of DSmT for Information Fusion'', Vol.3 (in preparation), 2008.

\bibitem{Smets90}
Ph.~Smets, ``Constructing the pignistic probability function in a context of uncertainty'', \emph{Uncertainty in AI}, vol.~5, pp. 29-39, 1990.

\bibitem{Smets2005}
Ph.~Smets, ``Decision making in the TBM: the necessity of the pignistic transformation", \emph{Int. Jour. Approx. Reasoning}, vol. 38, 2005.

\bibitem{Smets90b}
Ph.~Smets, ``The Combination of Evidence in the Transferable Belief Model'', \emph{IEEE Trans. on PAMI}, vol.~12, no.~5, pp. 447-458, 1990.

\bibitem{Sudano2001}
J.~Sudano, ``Pignistic Probability Transforms for Mixes of Low- and High-Probability Events'', \emph{Proc. of Fusion 2001}, Montreal, August 2001.


\bibitem{Sudano2002}
J.~Sudano, ``The system probability information content (PIC) \ldots '', \emph{Proc. of Fusion 2002}, Annapolis, July 2002.

\bibitem{Sudano2003}
J.~Sudano, ``Equivalence Between Belief Theories and Naive Bayesian Fusion for Systems with Independent Evidential Data - Part I, The Theory'', \emph{Proc. of Fusion 2003 }, Cairns, July 2003.

\bibitem{Sudano2006} 
J.~Sudano, ``Yet Another Paradigm Illustrating Evidence Fusion (YAPIEF)'', \emph{Proc. of Fusion 2006}, Florence, July 2006.

\end{thebibliography}

\end{document}